\documentclass[10pt,twocolumn,letterpaper]{article}

\usepackage{cvpr}              

\usepackage{graphicx}
\usepackage{amsmath}
\usepackage{amssymb}
\usepackage{booktabs}
\usepackage[utf8]{inputenc} 
\usepackage[T1]{fontenc}    
\usepackage{color,xcolor}
\usepackage{epsfig}
\usepackage{graphicx}
\usepackage{adjustbox}
\usepackage{array}
\usepackage{booktabs}
\usepackage{colortbl}
\usepackage{float,wrapfig}
\usepackage{hhline}
\usepackage{multirow}
\usepackage{subcaption} %
\usepackage{amsmath,amsfonts,amsthm,amssymb}
\usepackage{bm}
\usepackage{nicefrac}
\usepackage{microtype}
\usepackage{changepage}
\usepackage{extramarks}
\usepackage{fancyhdr}
\usepackage{lastpage}
\usepackage{setspace}
\usepackage{soul}
\usepackage{xspace}
\usepackage{pifont}
\usepackage{cuted}
\usepackage{wrapfig}
\usepackage{url}
\usepackage{algorithm, algorithmic}
\usepackage{enumitem}
\usepackage{wasysym}
\usepackage{hyperref}
\hypersetup{colorlinks = true,
            linkcolor = blue,
            urlcolor  = blue,
            citecolor = blue,
            anchorcolor = blue}

\usepackage[capitalize]{cleveref}
\crefname{section}{Sec.}{Secs.}
\Crefname{section}{Section}{Sections}
\Crefname{table}{Table}{Tables}
\crefname{table}{Tab.}{Tabs.}

\def\cvprPaperID{3768} 
\def\confName{CVPR}
\def\confYear{2022}

\begin{document}

\newcommand{\xpar}[1]{\noindent\textbf{#1}\ \ }
\newcommand{\vpar}[1]{\vspace{3mm}\noindent\textbf{#1}\ \ }

\newcommand{\sect}[1]{Section~\ref{#1}}
\newcommand{\sects}[1]{Sections~\ref{#1}}
\newcommand{\eqn}[1]{Equation~\ref{#1}}
\newcommand{\eqns}[1]{Equations~\ref{#1}}
\newcommand{\fig}[1]{Figure~\ref{#1}}
\newcommand{\figs}[1]{Figures~\ref{#1}}
\newcommand{\tab}[1]{Table~\ref{#1}}
\newcommand{\tabs}[1]{Tables~\ref{#1}}
\newcommand{\x}{\mathbf{x}}
\newcommand{\y}{\mathbf{y}}
\newcommand{\fid}{Fr\'echet Inception Distance\xspace}
\newcommand{\lblfig}[1]{\label{fig:#1}}
\newcommand{\lblsec}[1]{\label{sec:#1}}
\newcommand{\lbleq}[1]{\label{eq:#1}}
\newcommand{\lbltbl}[1]{\label{tbl:#1}}
\newcommand{\lblalg}[1]{\label{alg:#1}}
\newcommand{\lblline}[1]{\label{line:#1}}

\newcommand{\ignorethis}[1]{}
\newcommand{\norm}[1]{\lVert#1\rVert_1}
\newcommand{\fcseven}{$\mbox{fc}_7$}

\renewcommand*{\thefootnote}{\fnsymbol{footnote}}

\def\naive{na\"{\i}ve\xspace}
\def\Naive{Na\"{\i}ve\xspace}

\makeatletter
\DeclareRobustCommand\onedot{\futurelet\@let@token\@onedot}
\def\@onedot{\ifx\@let@token.\else.\null\fi\xspace}

\def\iid{\emph{i.i.d}\onedot}
\def\eg{\emph{e.g}\onedot} \def\Eg{\emph{E.g}\onedot}
\def\ie{\emph{i.e}\onedot} \def\Ie{\emph{I.e}\onedot}
\def\cf{\emph{c.f}\onedot} \def\Cf{\emph{C.f}\onedot}
\def\etc{\emph{etc}\onedot} \def\vs{\emph{vs}\onedot}
\def\wrt{w.r.t\onedot} \def\dof{d.o.f\onedot}
\def\etal{\emph{et al}\onedot}
\def\vs{\textbf{\emph{vs}\onedot}}
\makeatother

\definecolor{citecolor}{RGB}{34,139,34}
\definecolor{mydarkblue}{rgb}{0,0.08,1}
\definecolor{mydarkgreen}{rgb}{0.02,0.6,0.02}
\definecolor{mydarkred}{rgb}{0.8,0.02,0.02}
\definecolor{mydarkorange}{rgb}{0.40,0.2,0.02}
\definecolor{mypurple}{RGB}{111,0,255}
\definecolor{myred}{rgb}{1.0,0.0,0.0}
\definecolor{mygold}{rgb}{0.75,0.6,0.12}
\definecolor{mydarkgray}{rgb}{0.66, 0.66, 0.66}

\newcommand{\supplement}[1]{\color{blue}{#1}}
\newcommand{\muyang}[1]{{\color{orange}{[}Muyang: #1{]}}}
\newcommand{\han}[1]{{\color{mydarkblue}{[}Han: #1{]}}}
\newcommand{\yihan}[1]{{\color{purple}{[}Yihan: #1{]}}}
\newcommand{\SH}[1]{{\color{red}{[}SH: #1{]}}}
\newcommand{\myparagraph}[1]{\paragraph{#1}}

\def\multirowcenter{-0.5\dimexpr \aboverulesep + \belowrulesep + \cmidrulewidth}
\renewcommand{\thefootnote}{\number\value{footnote}}

\newcommand\blfootnotetext[1]{%
  \begingroup
  \renewcommand\thefootnote{}\footnotetext{#1}%
  \endgroup
}

\title{Lite Pose: Efficient Architecture Design for 2D Human Pose Estimation}

\author{
Yihan Wang$^{1*}$
\qquad 
Muyang Li$^2$
\qquad
Han Cai$^3$
\qquad
Weiming Chen$^3$
\qquad
Song Han$^3$ \\
\small{
$^1$Tsinghua University \quad $^2$Carnegie Mellon University \quad $^3$Massachusetts Institute of Technology }\\
\url{https://litepose.mit.edu}
}
\maketitle

\blfootnotetext{$^*$ work done while interning at MIT HAN Lab}

\begin{abstract}
Pose estimation plays a critical role in human-centered vision applications. However, it is difficult to deploy state-of-the-art HRNet-based pose estimation models~\cite{cheng2020higherhrnet} on resource-constrained edge devices due to the high computational cost (more than 150 GMACs per frame). In this paper, we study efficient architecture design for real-time multi-person pose estimation on edge. We reveal that HRNet's high-resolution branches are redundant for models at the low-computation region via our \textbf{gradual shrinking} experiments. Removing them improves both efficiency and performance. Inspired by this finding, we design \textbf{LitePose}, an efficient single-branch architecture for pose estimation, and introduce two simple approaches to enhance the capacity of LitePose, including 
\textbf{Fusion Deconv Head} and \textbf{Large Kernel Convs}. Fusion Deconv Head  removes the redundancy in high-resolution branches, allowing scale-aware feature fusion with low overhead. Large Kernel Convs significantly improve the model's capacity and receptive field while maintaining a low computational cost. With only 25\% computation increment, $7\times7$ kernels achieve $+14.0$ mAP better than $3\times 3$ kernels on the CrowdPose dataset. On mobile platforms, LitePose reduces the latency by up to $5.0\times$ without sacrificing performance, compared with prior state-of-the-art efficient pose estimation models, pushing the frontier of real-time multi-person pose estimation on edge. 
Our code and pre-trained models are released at \url{https://github.com/mit-han-lab/litepose}.
\end{abstract}
\section{Introduction}

\begin{figure}
\centering
	\includegraphics[width=\linewidth]{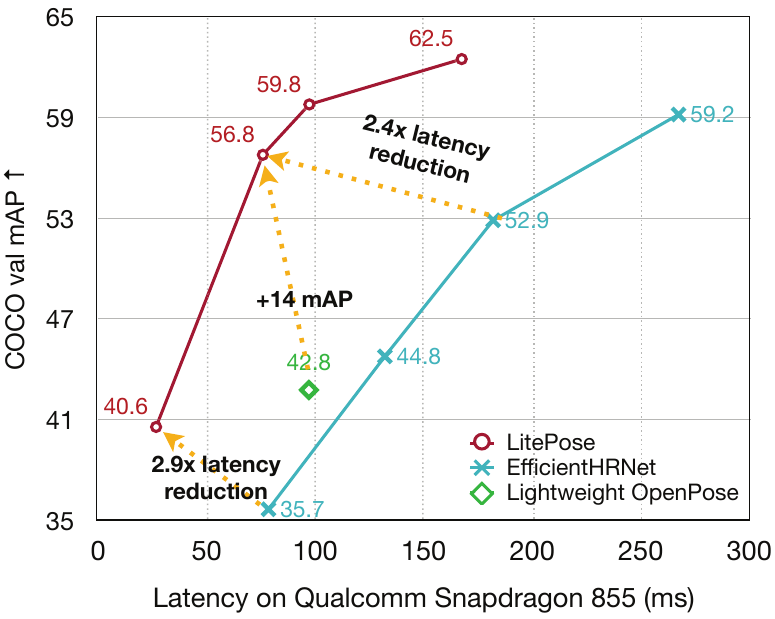}
\caption{LitePose provides up to $2.9\times$ latency reduction compared to EfficientHRNet~\cite{neff2020efficienthrnet} on Qualcomm Snapdragon 855 while achieving higher mAP on COCO. Compared with Lightweight OpenPose~\cite{osokin2018real}, LitePose obtains $14\%$ higher mAP on COCO with lower latency.}
\label{fig:coco}
\end{figure}

Human pose estimation aims to predict each person's keypoint positions from an image. It is a critical technique for many vision applications that require understanding human behavior. Typical human pose estimation models can be categorized into two paradigms: top-down and bottom-up. The top-down paradigm~\cite{kendall2015posenet,fang2017rmpe,he2017mask,wei2016convolutional,newell2016stacked,xiao2018simple,chen2018cascaded,sun2019deep} first detects people via an extra person detector and then performs single-person pose estimation for each detected person. In contrast, the bottom-up paradigm~\cite{newell2016stacked,pishchulin2016deepcut,insafutdinov2016deepercut,cao2019openpose,kreiss2019pifpaf,newell2016associative,cheng2020higherhrnet,geng2021bottom,jin2020differentiable,papandreou2018personlab} first predicts identity-free keypoints and then groups them into persons. As the bottom-up paradigm does not involve an extra person detector and does not require repeatedly running the pose estimation model for each person in the image, it is more suitable for real-time multi-person pose estimation on edge. 

However, existing bottom-up pose estimation models~\cite{newell2016stacked,pishchulin2016deepcut,insafutdinov2016deepercut,cao2019openpose,kreiss2019pifpaf,newell2016associative,cheng2020higherhrnet,geng2021bottom,jin2020differentiable,papandreou2018personlab} mainly focus on the high-computation region. For instance, HigherHRNet~\cite{cheng2020higherhrnet} achieves its best performance on the CrowdPose dataset~\cite{li2019crowdpose} with more than 150GMACs, which is prohibitive for edge devices. It is of great importance to design models with low computational cost while maintaining good performances. 

\begin{figure*}
\centering
	\includegraphics[width=\linewidth]{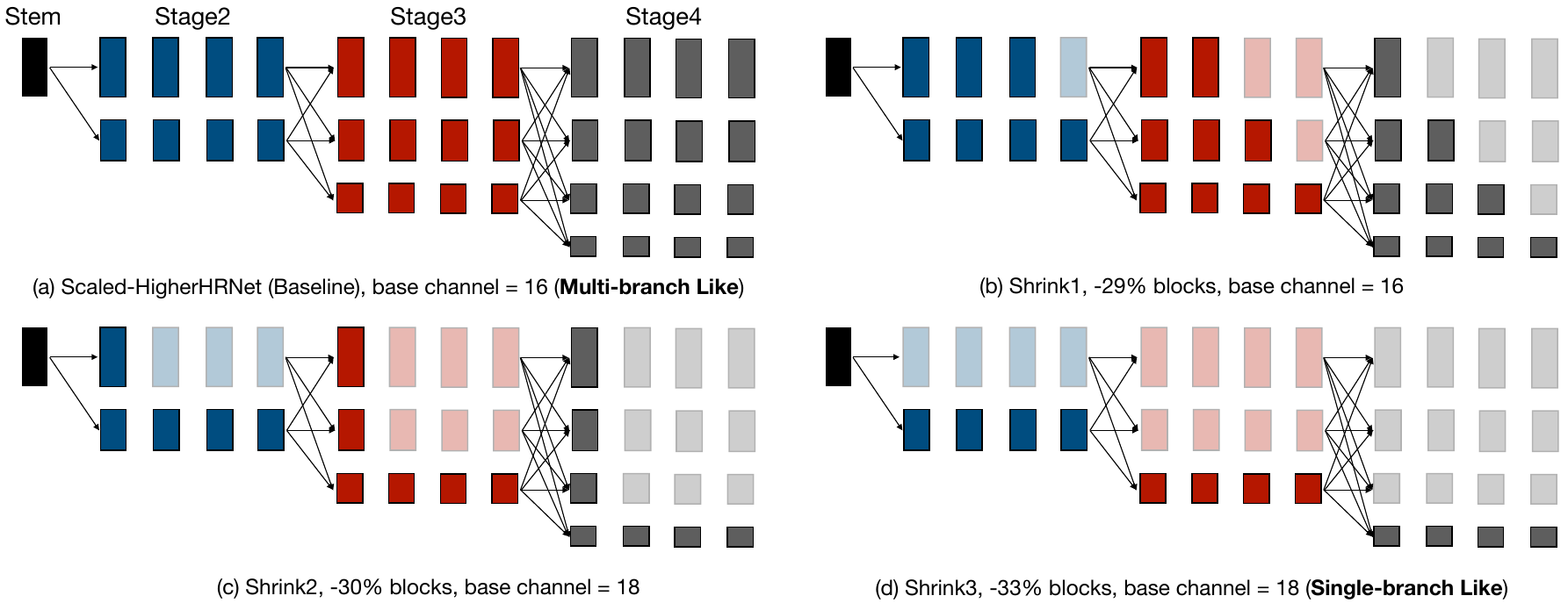}
\caption{Four architecture configurations in the \textbf{gradual shrinking} experiment. We use \textit{Scaled-HigherHRNet}\protect\footnotemark as the baseline for comparison. Removed blocks are shown in transparent. The network becomes increasingly close to the single-branch architecture from \textit{Baseline} to \textit{Shrink3}. To ensure different architecture configurations have similar MACs, we increase the base channel from 16 to 18 for \textit{Shrink2} and \textit{Shrink3}.}
\label{fig:3-shrink}
\end{figure*}

\begin{figure}
\centering
	\includegraphics[width=\linewidth]{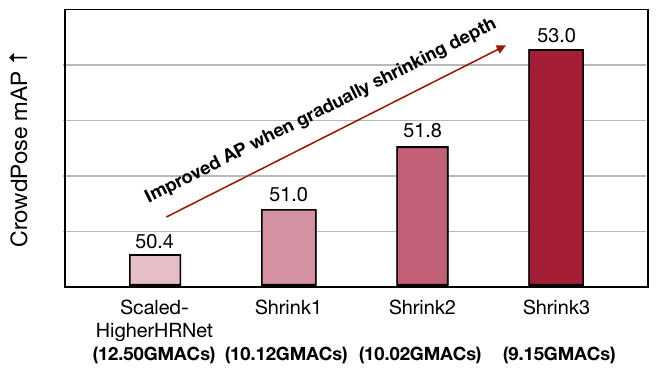}
\caption{The performance improves as we gradually shrink the high-resolution branches of \textit{Scaled-HigherHRNet-W16}.}
\label{fig:3-increase}
\end{figure}

In this paper, we study efficient architecture design for bottom-up human pose estimation. Previous study~\cite{cheng2020higherhrnet,geng2021bottom} in the high-computation region suggests that maintaining the high-resolution representation plays a critical role in achieving good performances for bottom-up pose estimation. However, it is unclear whether this still holds for models in the low-computation region. To answer this question, we build a ``bridge'' between the representative multi-branch architecture, HigherHRNet~\cite{cheng2020higherhrnet}, and the single-branch architecture by \textbf{gradual shrinking} (Figure~\ref{fig:3-shrink}). We surprisingly find that the performance improves as we shrink the depth of high-resolution branches for models in the low-computation region (Figure~\ref{fig:3-increase}). Inspired by this finding, we design a single-branch architecture, \textit{LitePose}, for efficient bottom-up pose estimation. In \textit{LitePose}, we use a modified MobileNetV2\footnote{Our method can also be combined with other backbones. We choose MobileNetV2 as it only contains basic operations (1$\times$1 conv, depthwise conv, Relu6) that are well-supported on most edge platforms.}~\cite{sandler2018mobilenetv2} backbone with two important improvement to efficiently handle the scale variation problem in the single-branch design: \textbf{fusion deconv head} and \textbf{large kernel conv}. The fusion deconv head removes the redundant refinement in high-resolution branches and therefore allows scale-aware multi-resolution fusion in a single-branch way (Figure~\ref{fig:4-fusion}). Meanwhile, different from image classification, we find large kernel convs provide a much more prominent improvement in bottom-up pose estimation (Figure~\ref{fig:4-kernel-size}). Finally, we apply Neural Architecture Search (NAS) to optimize the model architecture and choose appropriate input resolution.

Extensive experiments on CrowdPose~\cite{li2019crowdpose} and COCO~\cite{lin2014microsoft} demonstrate the effectiveness of \textit{LitePose}. On CrowdPose~\cite{li2019crowdpose}, \textit{LitePose} achieves $2.8\times$ MACs reduction and up to $5.0\times$ latency reduction with better performance. On COCO~\cite{lin2014microsoft}, LitePose obtains $2.9\times$ latency reduction compared with EfficientHRNet~\cite{neff2020efficienthrnet} while providing better performances. 

\noindent

We summarize our contributions as follows:
\begin{enumerate}
    \item We design \textbf{gradual shrinking} experiments, revealing that the high-resolution branches are redundant for models in the low-computation region.
    \item We propose \textit{LitePose}, an efficient architecture for bottom-up pose estimation. We also introduce two techniques to enhance the capacity of LitePose, including \textbf{fusion deconv head} and \textbf{large kernel conv}.
    \item Extensive experiments on two benchmark datasets, Microsoft COCO~\cite{lin2014microsoft} and CrowdPose~\cite{li2019crowdpose}  demonstrate the effectiveness of our method: LitePose achieves up to $2.8\times$ MACs reduction and up to $5.0\times$ latency reduction compared with state-of-the-art HRNet-based models.
\end{enumerate}

\section{Related Work}

\myparagraph{2D Human Pose Estimation.}

2D human pose estimation aims at localizing human anatomical keypoints (\eg, elbow, wrist) or parts. There are two main frameworks: the top-down framework and the bottom-up framework. Top-down methods~\cite{kendall2015posenet,fang2017rmpe,he2017mask,wei2016convolutional,newell2016stacked,xiao2018simple,chen2018cascaded,sun2019deep} perform single-person pose estimation by firstly detecting each person from the image. On the contrary, bottom-up methods~\cite{newell2016stacked,pishchulin2016deepcut,insafutdinov2016deepercut,cao2019openpose,kreiss2019pifpaf,newell2016associative,cheng2020higherhrnet,geng2021bottom,jin2020differentiable,papandreou2018personlab} directly predict keypoints of each person in an end-to-end manner. Typical bottom-up methods consist of two steps: predicting keypoint heatmaps and then grouping the detected keypoints into persons. Among these approaches, HRNet-based multi-branch architectures~\cite{cheng2020higherhrnet,geng2021bottom} provide state-of-the-art results. They design a multi-branch architecture to allow multi-resolution fusion, which has been proven effective in solving scale variation problems for bottom-up pose estimation. However, all these approaches are too computationally intensive (most >150GMACs) to be deployed on edge devices. In this work, we focus on the bottom-up framework for efficiency. Following state-of-the-art HRNet-based approaches~\cite{cheng2020higherhrnet}, we use associative embedding~\cite{newell2016associative} for grouping.

\myparagraph{Model Acceleration.} Apart from designing efficient models directly~\cite{howard2017mobilenets,sandler2018mobilenetv2,ma2018shufflenet,zhang2018shufflenet,liu2019point,wu2020lite}, another approach for model acceleration is to compress existing large models. Some methods aim at pruning the redundancy inside connections and convolution filters~\cite{han2015deep,han2015learning,wen2016learning,he2017channel,lin2017runtime,liu2017learning}. Meanwhile, some other methods focus on quantizing the network~\cite{courbariaux2016binarized,zhu2016trained,krishnamoorthi2018quantizing,wang2019haq}. Besides, several AutoML methods have also been proposed to automate the model compression and acceleration~\cite{he2018amc,yang2018netadapt,wang2019haq,liu2019metapruning}.
Recently, Yu \etal design LiteHRNet~\cite{yu2021lite} for top-down pose estimation, while we focus on the bottom-up paradigm. Neff \etal propose EfficientHRNet~\cite{neff2020efficienthrnet} for the efficient bottom-up pose estimation. They apply the compound scaling idea in EfficientNet~\cite{tan2019efficientnet} to HigherHRNet~\cite{cheng2020higherhrnet} and achieve $1.5\times$ MACs reduction. However, their method still faces drastic performance degradation when the computational constraint becomes tighter. In this work, we push the MACs reduction ratio to $5.1\times$ and achieves up to $5.0\times$ latency reduction on mobile platforms compared to EfficientHRNet. 

\myparagraph{Neural Architecture Search.}

Neural Architecture Search~(NAS) has achieved great success on large-scale image classification tasks~\cite{liu2017hierarchical,liu2018progressive,zoph2016neural,cai2018efficient}. Automatically designed models significantly outperform hand-crafted ones. To make the search process more efficient, researchers proposed one-shot NAS methods~\cite{liu2018darts,cai2018proxylessnas,wu2019fbnet,howard2019searching,guo2020single,bender2018understanding,cai2019once} in which different sub-networks share the same set of weights. To further explore the potential of our proposed architecture, we apply the \textit{once-for-all}~\cite{cai2019once} approach to automatically prune the redundancy inside channels and select the appropriate input size. Compared to the manually designed models trained from scratch, our searched models achieve prominent up to $+3.6$AP improvement.

\footnotetext{Scaled-HigherHRNet denotes the scale-down version of HigherHRNet.}
\section{Rethinking the Efficient Design Space}

\begin{figure*}[t]
\centering
	\includegraphics[width=\linewidth]{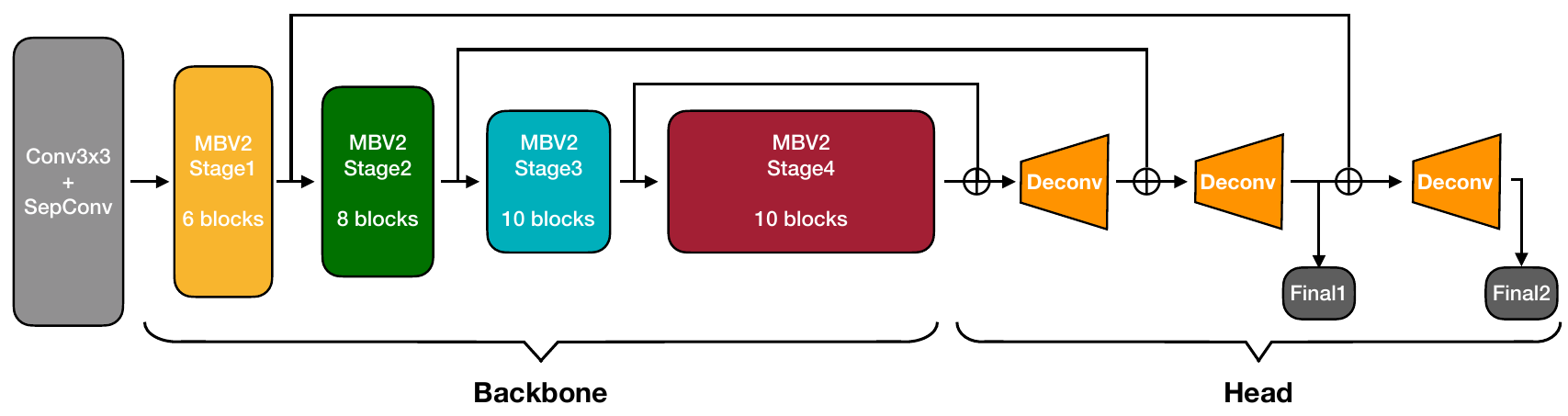}
\caption{The architecture of \textit{LitePose}. \textit{LitePose} consists of the backbone and the fusion deconvolution head. $\bigoplus$ means the ``concatenate" operation. The final convs are used for multi-resolution supervision following~\cite{cheng2020higherhrnet}.}
\label{fig:4-arch}
\end{figure*}

Multi-branch networks have achieved great success on the bottom-up pose estimation task. Their representative, HigherHRNet~\cite{cheng2020higherhrnet}, uses multi-branch architecture to help fuse multi-resolution features, which significantly alleviate the scale variation problem. Benefiting from this, multi-branch architectures outperform single-branch architectures and obtain state-of-the-art results. But there remains a problem in most of these methods~\cite{cheng2020higherhrnet,geng2021bottom,papandreou2018personlab,newell2016stacked} that they achieve their best performance with more than 150GMACs. The comparisons among methods are also mostly conducted with such high computation. Towards real-world edge applications, studies on efficient human pose estimation with lower computation are of high priority. In this section, we first introduce HRNet-based multi-branch architectures and how they cope with the scale variation problem. Then we point out the redundancy in high-resolution branches by \textbf{gradual shrinking} in computationally limited cases. Based on this observation, we propose the \textbf{fusion deconv head}, which removes the redundant refinement in high-resolution branches and therefore handles the scale variation problem in an efficient way. On the other hand, we empirically find the large kernels provide much more prominent improvement on the pose estimation task compared with the image classification task. Extensive experiments and ablation studies show the effectiveness of our method and reveal a fact that properly designed single-branch architectures can achieve better performance and lower latency. 

\subsection{Scale-Aware Multi-branch Architectures}

\myparagraph{Scale-Awareness.} The multi-branch design aims to alleviate the scale variation problem in bottom-up pose estimation. Since we need to predict the joint coordinates of all persons in an image, it is usually hard for single-branch architectures to recognize small persons and distinguish close joints from final low-resolution features, as shown in Figure~\ref{fig:visualization}(b). The high-resolution features introduced by multi-branch architectures, however, can reserve more detailed information and therefore help neural networks better capture small persons and discriminate close joints.

\myparagraph{Mechanism.} As shown in Figure~\ref{fig:3-shrink}, the main part of HRNet-based multi-branch architecture~\cite{cheng2020higherhrnet,geng2021bottom} consists of 4 stages. In stage $n$ (we regard stem as stage $1$ here), there are $n$ branches handling $n$ different input feature maps with different resolutions, respectively. When processing input features, each branch first refines its own input feature respectively, then exchanges information among branches to obtain multi-scale information.

\subsection{Redundancy in High-Resolution Branches}

However, when focusing on the performance with lower computation, we find the multi-branch architecture may not be the most efficient choice. In this section, we propose a method called \textbf{gradual shrinking} to reveal the redundancy in the high-resolution branches of the multi-branch architecture. As shown in Figure~\ref{fig:3-shrink} and Figure~\ref{fig:3-increase}, by gradually shrinking the depth of high-resolution branches, the multi-branch network behaves increasingly like a single-branch network. However, the performance does not degrade even improve.

\myparagraph{Gradual Shrinking.} To reveal the redundancy inside the HRNet-based multi-branch architecture~\cite{cheng2020higherhrnet,geng2021bottom}, we design a gradual shrinking experiment on the branches in each stage. Let $A_n=[a_{1}, \ldots, a_{n}]$ denote the number of blocks used to refine features for each branch ($a_i$ stands for the number of blocks in branch $i$) in stage $n$ before fusion. Here, branch $i$ processes feature maps with higher resolution than branch $i+1$. Then we can define the configuration of the whole multi-branch architecture as $A=\{A_{1},A_{2},A_{3},A_{4}\}$. We say $A'_{i}=[a'_{1},\ldots,a'_{i}]$ is shrunk from $A_{i}=[a_{1},\dots,a_{i}]$ if $\forall j\in\{1,\ldots,i\}$, $a'_{j}\le a_{j}$. For convenience, we denote this as $A'_{i} \le A_{i}$. A configuration $A'$ is said to be shrunk from $A$ (\ie, $A'\le A$) if $\forall i\in\{1,2,3,4\}$, $A'_{i}\le A_{i}$. With the aforementioned notations, \textbf{gradual shrinking} means that we construct a sequence of configurations $[C_{1}, \ldots, C_{m}]$ s.t. $C_{i+1}\le C_{i}, \forall i\in\{1,\ldots,m-1\}$. As shown in Figure~\ref{fig:3-shrink} and Figure~\ref{fig:3-increase}, we gradually shrink the depth of high-resolution branches and surprisingly find that this shrinking operation even helps improve the performance. Meanwhile, the gradual shrinking process makes the whole network increasingly similar to a single-branch network, which provides strong evidence that the single-branch architecture is more suitable for the efficient architecture design on the bottom-up pose estimation task. To make the gradual shrinking process clearer, we list the four configurations we use in detail below:

\begin{itemize}
    \item Baseline: $C_{1}=\{[4],[4,4],[4,4,4],[4,4,4,4]\}$, 12.5GMACs, base channel=16
    \item Shrink 1: $C_{2}=\{[4],[3,4],[2,3,4],[1,2,3,4]\}$, 10.1GMACs, base channel=16
    \item Shrink 2: $C_{3}=\{[4],[1,4],[1,1,4],[1,1,1,4]\}$, 10.0GMACs, base channel=18
    \item Shrink 3: $C_{4}=\{[4],[0,4],[0,0,4],[0,0,0,4]\}$, \ \ 9.2GMACs, base channel=18
\end{itemize}

\subsection{Fusion Deconv Head: Remove the Redundancy}

\begin{figure}
\centering
	\includegraphics[width=\linewidth]{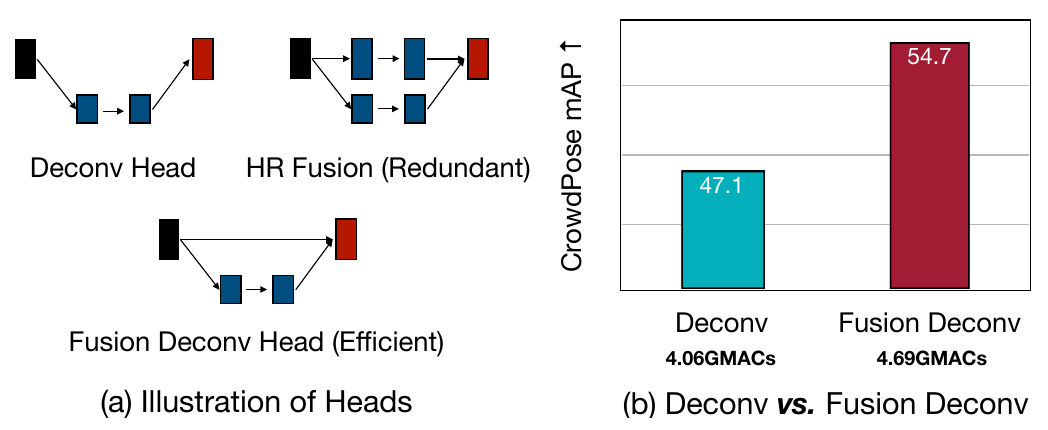}
\caption{Unlike conventional single-branch deconv head (from the black block to the red block), our fusion deconv head takes the advantage of HR fusion module and remove the high-resolution redundant refinement blocks. It achieves great improvement (+7.6AP) comparing to normal deconv head with minor computation increase.}
\label{fig:4-fusion}
\end{figure}

Though we have shown the redundancy in the multi-branch architecture above, its strong capability of handling the scale variation problem is still remarkable. Can we combine this feature into our design while keeping the merits of single-branch architecture (\eg, high efficiency)? To achieve this goal, we propose the fusion deconvolution layers as our final prediction head. To be specific, as shown in Figure~\ref{fig:4-arch} and \ref{fig:4-fusion}(b), we \textbf{directly} (\ie without any refinement) utilize the low-level high-resolution features generated by previous stages for deconvolution and final prediction layers. On the one hand, our \textit{LitePose} uses the single-branch architecture as our backbone, which benefits from the low-latency characteristic. On the other hand, directly using low-level high-resolution features avoids the redundant refinement in multi-branch HR fusion modules. Therefore, LitePose inherits the advantages from both single-branch design and multi-branch design in an efficient way. In Figure~\ref{fig:4-fusion}(a) and Figure~\ref{fig:visualization}, we show the strength of our fusion deconvolution head. With a negligible computational cost increase, we obtain a significant performance improvement (+7.6AP).

\subsection{Mobile Backbone with Large Kernel Convs}

Several papers~\cite{howard2017mobilenets,sandler2018mobilenetv2,ma2018shufflenet,zhang2018shufflenet} have studied efficient architectures under tight computational constraints on the image classification task. As shown in Figure~\ref{fig:4-arch}, we use a modified MobileNetV2~\cite{sandler2018mobilenetv2} architecture as the backbone in \textit{LitePose}. Following ~\cite{zhang2020efficientpose}, we make a minor modification on the original MobileNetV2~\cite{sandler2018mobilenetv2} backbone by removing the final down-sampling stage. Too many down-sampling layers will cause essential information loss, which is harmful to the high-resolution output of the pose estimation task.

\begin{figure}
\centering
	\includegraphics[width=\linewidth]{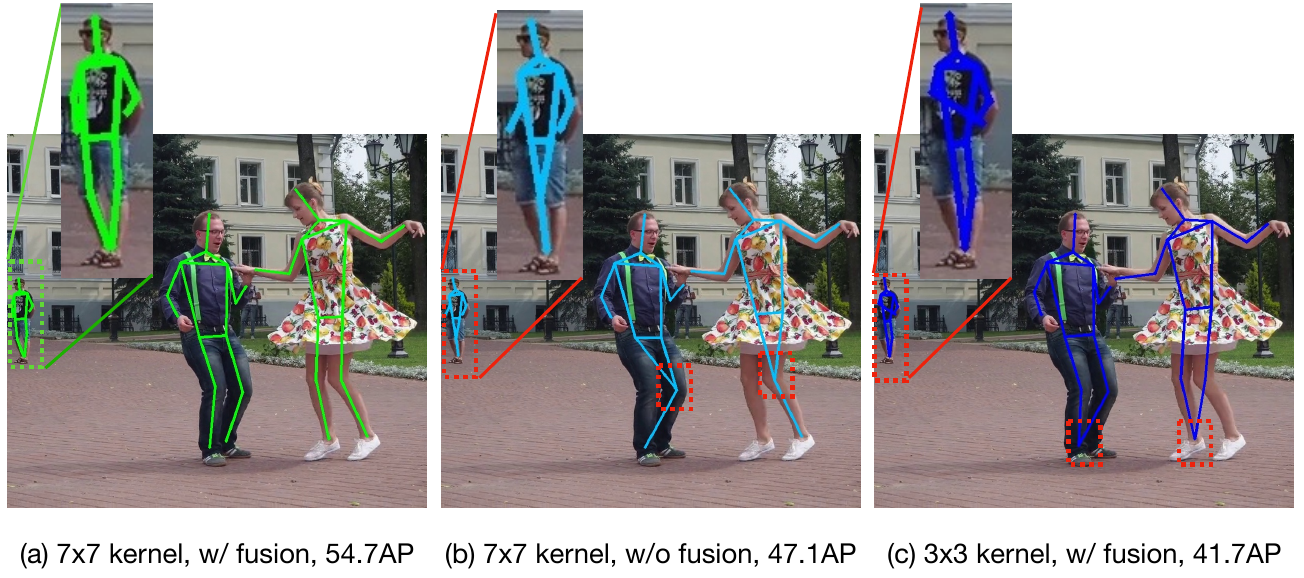}
\caption{Visualization of models with/without larger kernel convs and fusion deconv head. \textit{LitePose} can better recognize small persons and distinguish close joints with larger kernel convs and fusion deconv head.}
\label{fig:visualization}
\end{figure}

\begin{figure}
\centering
	\includegraphics[width=\linewidth]{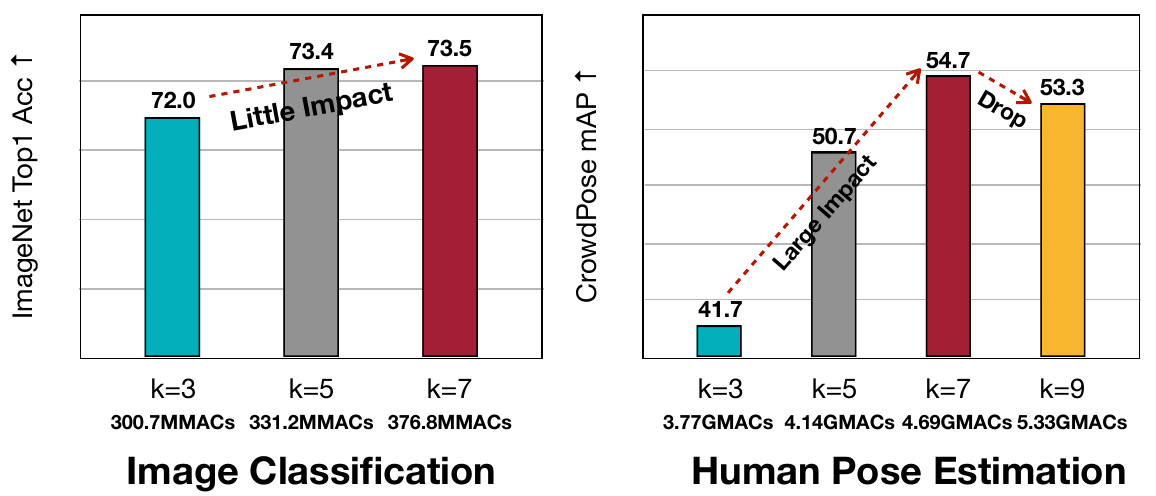}
\caption{$k$ represents the kernel size. Increasing the kernel size provides moderate performance improvement for image classification but making a big difference for pose estimation. Specifically, increasing the kernel size from 3 to 7 provides 13\% mAP improvement on CrowdPose.}
\label{fig:4-kernel-size}
\end{figure}

To further alleviate the scale variation problem, we introduce large kernels into our efficient architecture design. Unlike the traditional image classification task, this modification plays a much more important role in our proposed MobileNetV2-based~\cite{sandler2018mobilenetv2} backbone. In Figure~\ref{fig:4-kernel-size}, we show the performance comparisons among models with kernel sizes 3,5,7 (and 9 only for pose estimation) on both the image classification and the pose estimation task. With a similar computational cost increase (about +25\%), the performance gain on the pose estimation task (+13.0AP) is much more significant than on the image classification task (+1.5\% Acc). The visualization results in Figure~\ref{fig:visualization} also verify our claim. However, the rule is not ``the larger, the better''. 
Too large kernels will introduce many useless parameters and nonnegligible noise, which makes the training more difficult and incurs performance degradation, demonstrated in Figure~\ref{fig:4-kernel-size} for $k=9$ case.
Since we further find incorporating kernel size into the search space will severely degenerate the performance of NAS mentioned in Section~\ref{sec:4.3}, which may be caused by the large impact of the kernel size variation, we fix the kernel size to $7\times 7$ in our architecture.

\subsection{Single Branch, High Efficiency} 

\begin{figure}
\centering
	\includegraphics[width=\linewidth]{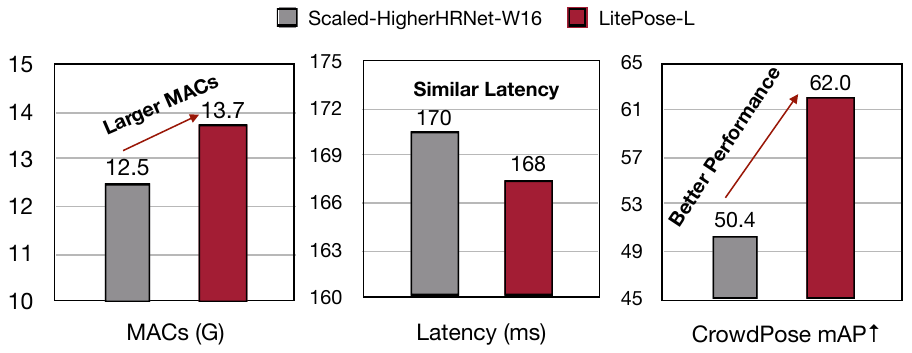}
\caption{Compared with the \textbf{multi-branch} \textit{Scaled-HigherHRNet-W16}~\cite{cheng2020higherhrnet}, \textbf{single-branch} \textit{LitePose-L} executes with higher parallelism. Therefore it achieves much better performance and similar latency on Qualcomm Snapdragon 855 with even higher MACs.} 
\label{fig:3-latency}
\end{figure}

Besides the performance, another important advantage of our single-branch \textit{LitePose} is its hardware-friendly characteristic. As mentioned in ShuffleNetV2~\cite{ma2018shufflenet}, network fragmentation such as multi-branch design reduces the degree of parallelism on some hardware. Therefore, towards real-world applications, single-branch architecture is a better choice. In summary, we show the quantitative comparison results between \textit{Scaled-HigherHRNet-W16}~\cite{cheng2020higherhrnet} and \textit{LitePose-L} in Figure~\ref{fig:3-increase}. Compared with \textit{Scaled-HigherHRNet-W16}~\cite{cheng2020higherhrnet}, \textit{LitePose-L} not only achieves much better performance ($+11.6$AP), but also obtains similar latency on Qualcomm Snapdragon 855  with even with larger MACs. All these results demonstrate the high efficiency of our single-branch \textit{LitePose}. 
\section{Neural Architecture Search}
\label{sec:4.3}

\begin{table*}[t]
\setlength{\tabcolsep}{3pt}
  \centering
  \small
  \begin{tabular}{lccccccccccccc}
    \toprule
    \multirow{2}{*}[\multirowcenter]{Model} & \multirow{2}{*}[\multirowcenter]{Input Size} & \multirow{2}{*}[\multirowcenter]{\#Params} & \multicolumn{2}{c}{\multirow{2}{*}[\multirowcenter]{\#MACs}} & \multicolumn{6}{c}{Latency (ms)} & \multirow{2}{*}[\multirowcenter]{AP} & \multirow{2}{*}[\multirowcenter]{AP$^{50}$} & \multirow{2}{*}[\multirowcenter]{AP$^{75}$}\\
    \cmidrule{6-11} 
    & & & & & \multicolumn{2}{c}{Nano} & \multicolumn{2}{c}{Mobile} & \multicolumn{2}{c}{Pi}\\
    \midrule
    HigherHRNet-W48~\cite{cheng2020higherhrnet} & 640$\times$640 & 63.8M & 154.6G & -- & 2101 & -- & 1532 & -- &12302& -- & 65.9 & 86.4 & 70.6\\
    \midrule
    Scaled-HigherHRNet-W24 & 512$\times$512 & 14.9M & 25.3G & -- & 330 & -- & 289 & -- & 1414 & -- & 57.4 & 83.2 & 63.2\\
    EfficientHRNet-H$_{-1}$~\cite{neff2020efficienthrnet} & 480$\times$480 & 13.0M & 14.2G & (1.8$\times$) & 283 & (1.2$\times$ ) & 267 & (1.1$\times$) & 1229 & (1.2$\times$) &56.3 & 81.3 &59.0\\
    \textbf{LitePose-S (Ours)} & 448$\times$448 & 2.7M & 5.0G & (5.1$\times$) & 97 & (3.4$\times$)& 76 & (3.8$\times$)&420 & (3.4$\times$)& 58.3 & 81.1 & 61.8\\
    \midrule
   Scaled-HigherHRNet-W16 & 512$\times$512 & 7.2M & 12.5G & -- & 172 & -- & 170 & -- & 898 & -- & 50.4 & 78.4 & 54.5\\
    EfficientHRNet-H$_{-3}$~\cite{neff2020efficienthrnet} & 416$\times$416 & 5.3M & 4.3G & (2.9$\times$) & 111 & (1.5$\times$) & 132 & (1.3$\times$) & 544 & (1.7$\times$) & 46.1 & 79.3 & 48.3\\
    \textbf{LitePose-XS (Ours)} & 256$\times$256 & 1.7M & 1.2G & (10.4$\times$) & 22 & (7.8$\times$) & 27 & (6.3$\times$) & 109 & (8.2$\times$) & 49.5 & 74.5 & 51.4\\
    \bottomrule
  \end{tabular}
    \caption{Results on CrowdPose \textit{test} set~\cite{li2019crowdpose}. Nano, Mobile, and Pi denote NVIDIA Jetson Nano GPU, Qualcomm Snapdragon 855, and Raspberry Pi 4B+, respectively.  \textit{LitePose} achieves better performance with up to $10.4\times$ MACs reduction and $8.2\times$ latency reduction on mobile platforms compared with Scaled-HigherHRNet-W16.}
  \label{tab:CrowdPose}
\end{table*}

Existing work~\cite{cheng2020higherhrnet,geng2021bottom,kreiss2019pifpaf,papandreou2018personlab,cao2019openpose,newell2016stacked,osokin2018real} on the bottom-up pose estimation task usually uses a hand-crafted (and mostly uniform) channel width across all the layers in the model and a fixed large resolution (\eg, $512\times512$). To further explore the potential compactness of our model, in this section, we apply \textit{once-for-all}~\cite{cai2019once} to automatically prune the redundancy in channels and select the optimal input resolution. The optimization goal and the search process are described in the following. Through NAS, we get four LitePose models (XS, S, M, and L) for different computation budgets. In Section~\ref{subsec:ablation}, we show the effectiveness of NAS in detail.


\myparagraph{Optimization Goal.} 
Suppose that the original LitePose architecture contains $\{c_k\}_{k=1}^K$ channels in each layer, where $K$ denotes the number of layers of the network. Our optimization goal is to find a sub-network whose input resolution is $r' < r$ with channel width $\{c'_k\}_{k=1}^K$ where $c'_k \le c_k$, such that it could meet our efficiency constraint while achieving the best Average Precision (AP).

\myparagraph{One-shot Supernet Training.} We first train a LitePose supernet that supports different channel number configurations via weight sharing following~\cite{cai2019once,guo2020single}. For each training iteration, we uniformly sample a channel configuration and train the supernet with it. 
In this way, each configuration is equally trained and could operate independently.
To help the supernet learn better associate embedding~\cite{newell2016associative} for grouping, we initialize the supernet with pre-trained weights. See Section~\ref{sec:exp-setting} for more details about the supernet training and pre-training.

\myparagraph{Search \& Fine-tune.} Since the supernet is thoroughly trained with weight sharing, we could directly extract the weights of a certain sub-network and evaluate the sub-network without further fine-tuning. This approximates the final performance of the sub-network. We use the evolutionary algorithm~\cite{real2019regularized} to find the optimal configurations given specific efficiency constraints (\eg, MACs). After finding optimal configurations, we fine-tune the corresponding sub-networks for several epochs and report the final performance. See Section~\ref{sec:5} for more details about the fine-tuning.

\section{Experiments}

\label{sec:5}
\begin{figure*}
\centering
	\includegraphics[width=\linewidth]{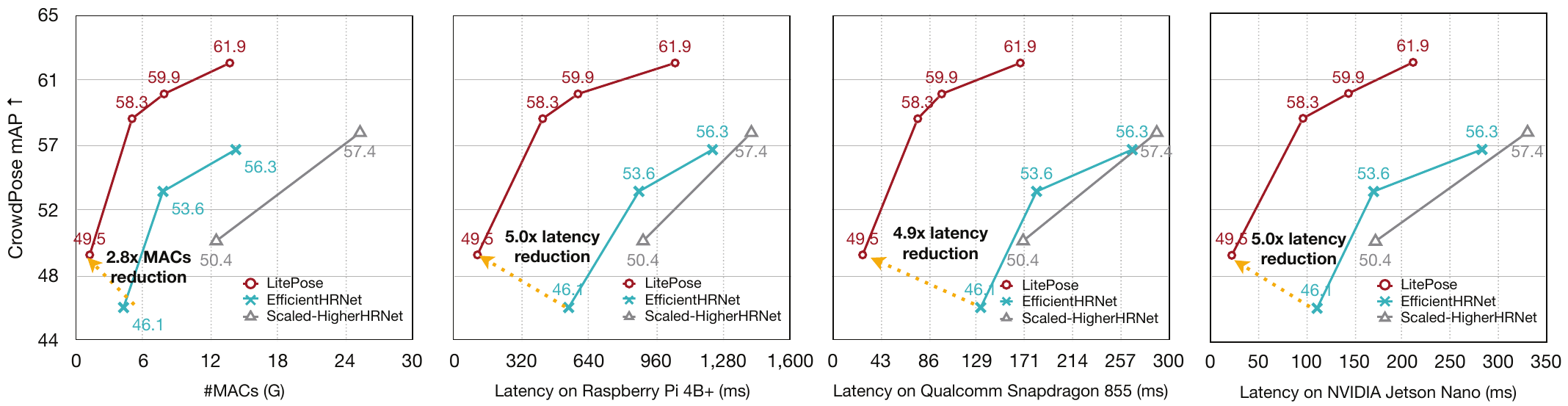}
\caption{On CrowdPose~\cite{li2019crowdpose}, LitePose achieves $2.8\times$ MACs reduction compared to EfficientHRNet~\cite{neff2020efficienthrnet}. The hardware-friendly design of LitePose allows high parallelism and therefore achieves much lower latency on various mobile platforms: It achieves $5.0\times$, $4.9\times$, and $5.0\times$ latency reduction on Raspberry Pi 4B+, Qualcomm Snapdragon 855, and NVIDIA Jetson Nano respectively.}
\label{fig:crowdpose}
\end{figure*}

\subsection{Dataset \& Metrics}

\myparagraph{Microsoft COCO.} Microsoft COCO~\cite{lin2014microsoft} contains over 200,000 images with 250,000 person instances labeled with 17 keypoints. It is divided into \textit{train/val/test-dev} sets with 57k, 5k, and 20k images, respectively. All our experiments on Microsoft COCO~\cite{lin2014microsoft} are trained only on the \textit{train} set. And we report the results on both \textit{val} and \textit{test-dev} sets.

\myparagraph{CrowdPose.} CrowdPose~\cite{li2019crowdpose} consists of 20,000 images, containing about 80,000 persons labeled with 14 keypoints. Compared to Microsoft COCO~\cite{lin2014microsoft}, CrowdPose~\cite{li2019crowdpose} contains more crowded scenes, posing more challenges to pose estimation methods. Following HigherHRNet~\cite{cheng2020higherhrnet}, we train our models on the \textit{train+val} set and report our results on the \textit{test} set.

\myparagraph{Evaluation Metric.} The standard evaluation metric is based on Object Keypoint Similarity (OKS): OKS$=\frac{\sum_{i}{exp(-d_{i}^2/2s^2k_{i}^2)\delta(v_{i}>0)}}{\sum_{i}{\delta(v_{i}>0)}}$. Here $d_{i}$ represents the Euclidean distance between a detected keypoint and its corresponding ground truth position. $v_{i}$ denotes the visibility flag of keypoint $i$. $s$ is the object scale, and $k_{i}$ is a per-keypoint constant that controls falloff. Based on OKS, we report the standard average precision (AP), AP$^{50}$, and AP$^{75}$ as the experiment results.

\subsection{Experiment Setting}

\label{sec:exp-setting}


\myparagraph{Data Augmentation.} Following \cite{cheng2020higherhrnet} and \cite{xiao2018simple}, the data augmentation includes random rotation $[-30^\circ, 30^\circ]$, random scale $[0.75, 1.5]$, random translation ($[-40, 40]$), and random flip.

\myparagraph{Pre-training Details.} We find that the network will learn low-quality Associative Embedding (AE)~\cite{newell2016associative} if we train our one-shot supernet from scratch. To address this issue, we resort to pre-training. To be specific, we train the largest supernet without the AE loss (\ie, only the heatmap loss) on the Microsoft COCO \textit{train} set~\cite{lin2014microsoft} for 100 epochs. Then we use it as the pre-trained model for further supernet training.

\myparagraph{Supernet Training Setting.} We conduct the one-shot NAS on the CrowdPose dataset~\cite{li2019crowdpose}. 
We train \textit{LitePose-L/M/S} and \textit{LitePose-XS} with different training hyper-parameters and search space. We train \textit{LitePose-L/M/S} supernet for 800 epochs with batch size 32 and \textit{LitePose-XS} supernet for 2400 epochs with batch size 128. In each training step, we uniformly sample an architecture configuration from the search space and train the supernet with it ($lr=0.001$ for $bs=32$, $lr=0.004$ for $bs=128$).

\myparagraph{Fine-tuning Setting.} On CrowdPose dataset~\cite{li2019crowdpose}, we fix the architecture configuration and tune the model for 200 epochs with batch size 32. The original learning rate is set to $10^{-3}$, and drops to $10^{-4}$ and $10^{-5}$ at the $50_{th}$ and the $180_{th}$ epoch, respectively (linearly increase~\cite{he2019bag} for $bs=128$ case). On COCO dataset~\cite{lin2014microsoft}, we take the supernet trained on CrowdPose~\cite{li2019crowdpose} as the pre-trained model for initialization. For each searched configuration, we train the corresponding model for $500$ epochs with batch size 32. The original learning rate is set to $10^{-3}$, dropped to $10^{-4}$ and $10^{-5}$ at the $350_{th}$ and the $480_{th}$ epoch, respectively.

\myparagraph{Search Details.}

\begin{table*}[t]
\setlength{\tabcolsep}{3pt}
  \centering
  \small
  \begin{tabular}{lcccccccccccc}
    \toprule
    \multirow{2}{*}[\multirowcenter]{Model} & \multirow{2}{*}[\multirowcenter]{Input Size} & \multirow{2}{*}[\multirowcenter]{\#Params} & \multicolumn{2}{c}{\multirow{2}{*}[\multirowcenter]{\#MACs}} & \multicolumn{6}{c}{Latency (ms)} & \multirow{2}{*}[\multirowcenter]{AP$_{val}$} & \multirow{2}{*}[\multirowcenter]{AP$_{test\text{-}dev}$} \\
    \cmidrule{6-11} 
    & & & & & \multicolumn{2}{c}{Nano} & \multicolumn{2}{c}{Mobile} & \multicolumn{2}{c}{Pi}\\
    \midrule
    PersonLab~\cite{papandreou2018personlab} & 1401$\times$1401 & 68.7M & 405.5G & -- & \multicolumn{2}{c}{--} & \multicolumn{2}{c}{--} & \multicolumn{2}{c}{--} & 66.5 & 66.5\\
    Hourglass~\cite{newell2016stacked} & 512$\times$512 & 277.8M & 206.9G& -- & \multicolumn{2}{c}{--} & \multicolumn{2}{c}{--} & \multicolumn{2}{c}{--} & --&56.6\\
    HigherHRNet-W48~\cite{cheng2020higherhrnet} & 640$\times$640 & 63.8M & 155.1G & -- & 2101 & -- & 1532 & -- & 12302 & -- & 69.9 & 68.4\\
    \midrule
    Lightweight OpenPose~\cite{osokin2018real} & 368$\times$368 & 4.1M & 9.0G & -- & 155 & -- & 97 & -- & 562 & -- & 42.8 & --\\
    EfficientHRNet-H$_{-2}$~\cite{neff2020efficienthrnet} & 448$\times$448 & 8.3M & 7.9G & (1.1$\times$) & 170 & (0.9$\times$) &182 & (0.5$\times$) & 878 & (0.6$\times$) & 52.9 & 52.8\\
    \textbf{LitePose-S (Ours)} & 448$\times$448 & 2.7M & 5.0G & (1.8$\times$) & 97 & (1.3$\times$) & 76 & (1.3$\times$) & 420 & (1.3$\times$) & 56.8 & 56.7\\
    \midrule
    EfficientHRNet-H$_{-4}$ & 384$\times$384 & 2.8M & 2.2G & -- & 50 & -- & 78 & -- & 273 & -- & 35.7 & 35.5\\
    \textbf{LitePose-XS (Ours)} & 256$\times$256 & 1.7M & 1.2G & (1.8$\times$) & 22 & (2.3$\times$) & 27 & (2.9$\times$) &109 & (2.5$\times$) & 40.6 & 37.8\\
    \bottomrule
  \end{tabular}
  \caption{Results on COCO \textit{val/test-dev} set~\cite{lin2014microsoft}. Compared with EfficientHRNet~\cite{neff2020efficienthrnet}, LitePose achieves $1.8\times$ MACs reduction and up to $2.9\times$ latency reduction while providing better performances. Compared with Lightweight OpenPose~\cite{osokin2018real}, it obtains much higher performance ($+14.0$AP) with lower latency.}
  \label{tab:COCO}
\end{table*}

We conduct NAS on the CrowdPose dataset~\cite{li2019crowdpose}. After obtaining the searched architectures, we directly generalize them to the COCO dataset~\cite{lin2014microsoft} and report their performance on both datasets. For \textit{LitePose-L/M/S} supernet training, we choose resolution from $[512, 448]$ and channel width ratio from $[1.0, 0.75, 0.5]$. For \textit{LitePose-XS} supernet training, we choose resolution from $[512, 448, 384, 320, 256]$ and channel width ratio from $[1.0, 0.75, 0.5, 0.25]$.

\myparagraph{Measurement Details.} We measure the latency of our models on Qualcomm Snapdragon 855 GPU, Raspberry Pi 4B+, and NVIDIA Jetson Nano GPU. For real-world edge deployment, 
it is crucial for DL models to efficiently integrate some optimized libraries and runtimes as their backends and generate the fastest possible executable.
Therefore, all the latency results we report on raspberry Pi 4B+ and NVIDIA Jetson Nano GPU are optimized by TVM AutoScheduler~\cite{chen2018tvm,zheng2020ansor}, which can help us better simulate the latency of real-world applications.

\subsection{Ablation Experiments}
\label{subsec:ablation}


\begin{table}[h]
\setlength{\tabcolsep}{3.5pt}
  \centering
  \small
  \begin{tabular}{lccccccc}
    \toprule
    \multirow{2}{*}[\multirowcenter]{Arch} & \multicolumn{4}{c}{Setting} & \multirow{2}{*}[\multirowcenter]{\#MACs} & \multirow{2}{*}[\multirowcenter]{AP} \\
    \cmidrule{2-5}
    & Knl. & Fsn. & Spnt. & Dstl. \\
    \midrule
    0.5 LitePose & $3 \times 3$ & $\checkmark$ & & & 3.8G & 41.7 \\
    0.5 LitePose & $5 \times 5$ & $\checkmark$ & & & 4.1G & 50.7 \\
    0.5 LitePose & $7 \times 7$ & & & & 4.1G & 47.1 \\
    0.5 LitePose & $7 \times 7$ & $\checkmark$ & & & 4.7G & 54.7 \\
    0.5 LitePose & $7 \times 7$ & $\checkmark$ & $\checkmark$ & & 4.7G & 56.1 \\
    \textbf{LitePose-S (Ours)} & $7 \times 7$ & $\checkmark$ & $\checkmark$ & & 5.0G & 58.3 \\
    \midrule
    LitePose-XS & $7 \times 7$ & $\checkmark$ &  & & 1.2G & 45.7 \\
    LitePose-XS & $7 \times 7$ & $\checkmark$ & $\checkmark$ & & 1.2G & 48.4 \\
    \textbf{LitePose-XS (Ours)} & $7 \times 7$ & $\checkmark$ & $\checkmark$ & $\checkmark$ & 1.2G & 49.5\\
    \bottomrule
  \end{tabular}
  \caption{Ablation study. \textbf{0.5 LitePose} means linearly scales each layer to the LitePose supernet to 50\% channels. \textbf{Knl.}: Kernel size; \textbf{Fsn}: Fusion deconv head; \textbf{Spnt.}: Supernet training; \textbf{Dstl}: Distillation. Large kernels and the fusion deconv head provide $+13.0$AP and $+7.6$AP, respectively. The supernet training benefits both two configurations by up to $+2.7$AP. The architecture search brings $+2.2$AP to the manually designed model. And the distillation brings $+1.1$AP to \textit{LitePose-XS}. }
  \label{tab:ablation}
\end{table}

\myparagraph{Large Kernels.} As shown in Table~\ref{tab:ablation} and Figure~\ref{fig:4-kernel-size}, with only minor computation increase, the $7\times 7$ kernels enhance the capability of coping with scale variation problem and therefore provides the best performance.

\myparagraph{Fusion Deconv Head.} Another way to handle the scale variation problem is multi-resolution fusion as the introduction of large resolution features can help better capture small persons. We quantitatively show the performance gain in Table~\ref{tab:ablation} and Figure~\ref{fig:4-fusion}: our efficient fusion deconv head improve the performance by $+7.6$AP on CrowdPose~\cite{li2019crowdpose} dataset with only minor computation increase.

\myparagraph{Neural Architecture Search.} Neural Architecture Search (NAS) benefits our method from two aspects: one-shot supernet training and architecture search with fine-tuning. As shown in Table~\ref{tab:ablation}, supernet training provides $+1.4$AP and $+2.7$AP on \textit{0.5 LitePose} and \textit{LitePose-XS}, respectively. Architecture search also offers $+2.2$AP on \textit{0.5 LitePose}. Besides, for \textit{LitePose-XS}, we use \textit{LitePose-S} as its teacher for heatmap loss in fine-tuning and obtain $+1.1$AP. 

\subsection{Main Results}

\myparagraph{Results on CrowdPose.} We first report the results on the CrowdPose dataset~\cite{li2019crowdpose}. Compared to Microsoft COCO~\cite{lin2014microsoft}, it contains more crowded scenes. The strong assumption of top-down methods that each person detection box only contains a single person in the center is hard to satisfy in crowded scenes. Therefore, several top-down methods~\cite{he2017mask,fang2017rmpe} that perform well on the COCO dataset~\cite{lin2014microsoft} fail on the CrowdPose dataset~\cite{li2019crowdpose}. We also achieve a much better performance-computation trade-off than the state-of-the-art bottom-up HRNet-based baselines~\cite{cheng2020higherhrnet,neff2020efficienthrnet}. As shown in Table~\ref{tab:CrowdPose} and Figure~\ref{fig:crowdpose}, our architecture achieves $2.8\times$ MACs reduction and up to $5.0\times$ latency reduction on mobile platforms compared with HRNet-based methods.

\myparagraph{Results on Microsoft COCO.} We also show the results on Microsoft COCO dataset~\cite{lin2014microsoft}. Our method outperforms HRNet-based approaches~\cite{cheng2020higherhrnet,neff2020efficienthrnet} by a large margin. As shown in Table~\ref{tab:COCO} and Figure~\ref{fig:coco}, our architecture achieves up to $2.4\times$ with even higher performance. Besides, compared with Lightweight Openpose~\cite{osokin2018real}, our method performs much better ($+14.0$AP) with lower latency on mobile platforms. 

\section{Conclusion \& Discussion}

In this paper, we studied the efficient architecture design for multi-person pose estimation on edge devices. We designed a gradual shrinking experiment to bridge the multi-branch and single-branch architecture. Our study shows that the high-resolution branches are redundant for models in the low-computation region. Inspired by this, we propose LitePose, an efficient architecture for pose estimation, which inherits the merits of both the single-branch and multi-branch architecture. We also observe an interesting phenomenon that large kernels work much better than small kernels on the human pose estimation task. Extensive experiments demonstrate the effectiveness and robustness of LitePose, paving the way to real-time human pose estimation for edge applications.

\section*{Acknowledgement}
We thank National Science Foundation, MIT-IBM Watson AI Lab, Ford, Hyundai and Intel for supporting this research. We thank Ji Lin, Yaoyao Ding and Lianmin Zheng for their helpful comments on the project. We also thank Shengyu Wang and Ruihan Gao for their valuable feedback on the manuscript.







\end{document}